\documentclass[sigconf, preprint, nonacm]{acmart}
\AtBeginDocument{%
  }

\setcopyright{none}
\renewcommand\footnotetextcopyrightpermission[1]{}

\usepackage[table]{xcolor}

\usepackage{multirow}
\usepackage{makecell}
\usepackage{array}
\usepackage{graphicx}
\usepackage{subcaption}
\usepackage{pifont}
\newcommand{\cmark}{\ding{51}} % ✓
\newcommand{\xmark}{\ding{55}} % ✗
\newcolumntype{C}[1]{>{\centering\arraybackslash}p{#1}}

\definecolor{humanbg}{RGB}{220,235,227} % 浅绿色 human 行
\newcommand{\best}[1]{\cellcolor{gray!15}\textbf{#1}}
\newcommand{\second}[1]{\underline{#1}}
\begin{document}

\title{EgoCoT-Bench: Benchmarking Grounded and Verifiable Operation-Centric Chain of Thought Reasoning for MLLMs}


\author{Yang Dai}
\authornote{Both authors contributed equally to this research.}
\email{yangdai@zju.edu.cn}
\affiliation{%
  \institution{Zhejiang University}
  \city{Hangzhou}
  \state{Zhejiang}
  \country{China}
}
\author{Dian Jiao}
\authornotemark[1]
\email{jd_dcd@zju.edu.cn}
\affiliation{%
  \institution{Zhejiang University}
  \city{Hangzhou}
  \state{Zhejiang}
  \country{China}
}
\author{Tianwei Lin}

\email{lintw@zju.edu.cn}
\affiliation{%
  \institution{Zhejiang University}
  \city{Hangzhou}
  \state{Zhejiang}
  \country{China}
}
\author{Wenqiao Zhang}
\authornote{Corresponding author.}

\email{wenqiaozhang@zju.edu.cn}
\affiliation{%
  \institution{Zhejiang University}
  \city{Hangzhou}
  \state{Zhejiang}
  \country{China}
}

% \affiliation{%

% \affiliation{%

% \affiliation{%

% \affiliation{%

% \affiliation{%

\renewcommand{\shortauthors}{Dai et al.}

\begin{abstract}
The rapid development of Multimodal Large Language Models (MLLMs) has led to growing interest in egocentric video understanding, specifically the ability for MLLMs to recognize fine-grained hand-object interactions, track object state changes over time, and reason about manipulative processes in dynamic environments from a first-person perspective. However, existing egocentric video benchmarks suffer from \textbf{limited grounded rationale evaluation}, offering limited support for fine-grained operation-centric reasoning and rarely examining whether model rationales are grounded in explicit spatio-temporal evidence. To address this gap, we introduce \textbf{EgoCoT-Bench}, a fine-grained egocentric benchmark for grounded and verifiable operation-centric reasoning with explicit step-by-step rationale annotations. Overall, EgoCoT-Bench comprises 3,172 verifiable QA pairs over 351 egocentric videos separated into four task groups for a total of 12 sub-task groups, encompassing perception and retrospection, anticipation, and high-level reasoning. The benchmark is constructed through a spatio-temporal scene graphs (STSG) guided generation framework and is further refined by human annotators to ensure correctness, egocentric relevance and fine-grained quality. Experimental results show continuing difficulties with egocentric fine-grained reasoning and further reveal that many multimodal models produce explanations that are answer-correct, but have evidence that is inconsistent with the answer. We hope EgoCoT-Bench can serve as a useful testbed for grounded and verifiable reasoning in egocentric video understanding. Project page and supplementary materials are available at: \url{https://dstardust.github.io/EgoCoT/}.
\end{abstract}

\begin{CCSXML}
<ccs2012>
   <concept>
       <concept_id>10010147.10010178.10010224.10010225.10010228</concept_id>
       <concept_desc>Computing methodologies~Activity recognition and understanding</concept_desc>
       <concept_significance>500</concept_significance>
       </concept>
 </ccs2012>
\end{CCSXML}

\ccsdesc[500]{Computing methodologies~Activity recognition and understanding}

\keywords{egocentric video understanding, benchmark, multimodal large language models, grounded reasoning, fine-grained reasoning, verifiable rationales}
\begin{teaserfigure}
\centering
  \includegraphics[width=0.96\textwidth]{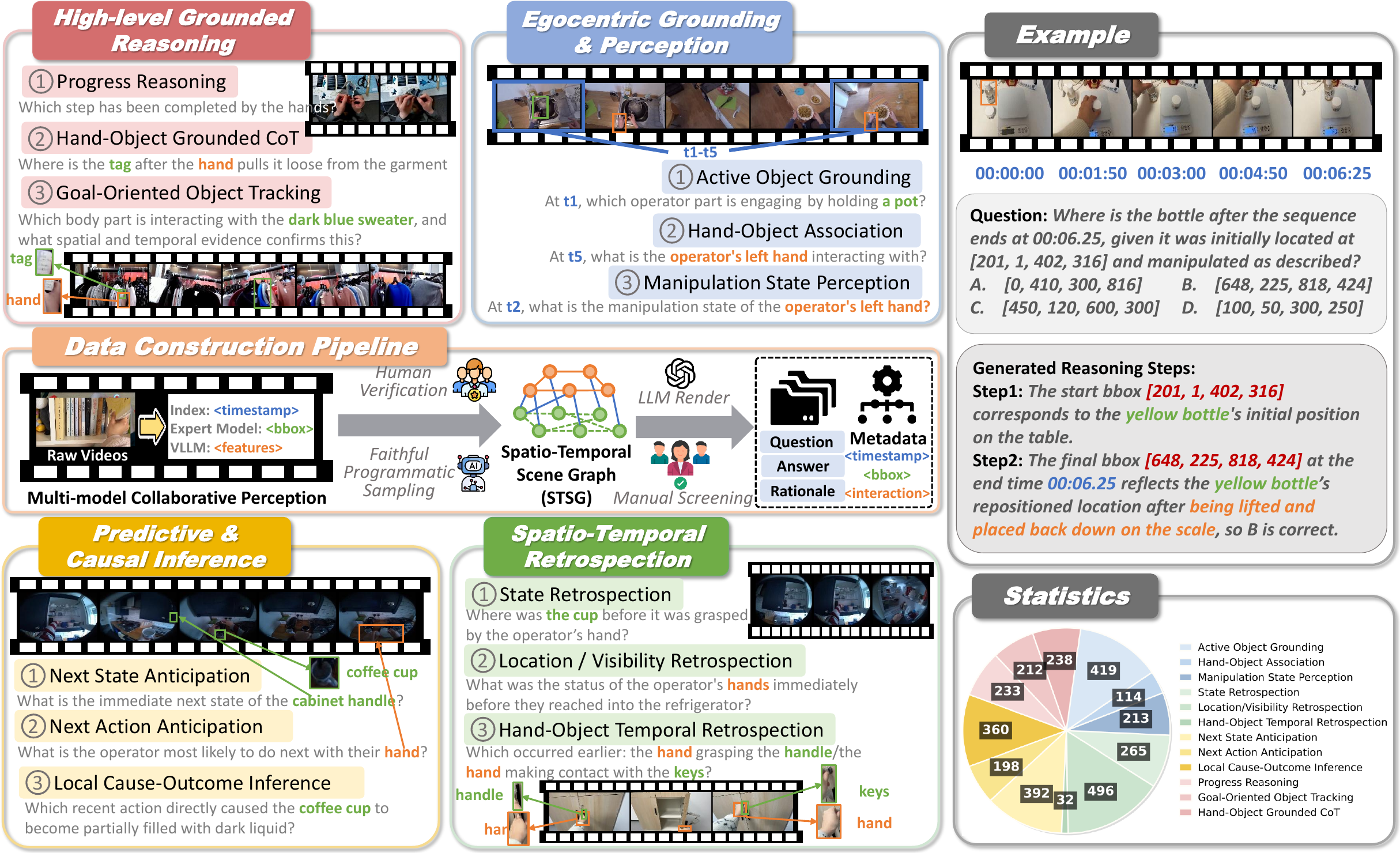}
    \caption{Overview of EgoCoT-Bench. EgoCoT-Bench is a fine-grained benchmark for grounded and verifiable operation-centric reasoning in egocentric videos, containing 3,172 QA pairs over 351 videos across four task groups and 12 subtasks. It is built through an STSG-guided human verification pipeline with explicit spatio-temporal evidence and rationale annotations.}
  % \Description{Enjoying the baseball game from the third-base
  % seats. Ichiro Suzuki preparing to bat.}
  \label{fig:overview}
\end{teaserfigure}


\maketitle

\section{Introduction}

The rapid progress of multimodal large language models (MLLMs) has greatly advanced video understanding, opening up new possibilities for question answering, temporal reasoning, and embodied perception~\cite{zhang-etal-2023-video, Maaz2023VideoChatGPT, vidllmsurvey, lin2025healthgpt, zhang2024hyperllava, zhong2026unified, yang2025graft}. Among these directions, egocentric video understanding is of particular importance for real-world assistive agents and embodied systems~\cite{Damen_2018_ECCV, Grauman_2022_CVPR, Majumdar_2024_CVPR}, since first-person observations directly capture how a user manipulates objects, shifts attention, and interacts with the surrounding environment during task execution. Compared with generic third-person videos, egocentric videos require models to reason about ongoing hand-object interactions, local state changes, and short-horizon action evolution from the operator's own viewpoint~\cite{Sener_2022_CVPR, Wang_2023_ICCV}.

\begin{table*}[t]
\caption{Comparison with representative video and egocentric benchmarks.}
\label{tab:benchmark_comparison}
\centering
\small
\setlength{\tabcolsep}{3.2pt}
\resizebox{0.98\textwidth}{!}{%
\renewcommand{\arraystretch}{1.00}
\begin{tabular}{lcc|ccc|cccc}
\toprule
\textbf{Benchmark} & \textbf{\#Clips} & \textbf{\#Samples} & \textbf{\makecell{Question\\Type}} & \textbf{Annotation} & \textbf{Egocentric} & \textbf{\makecell{CoT /\\Rationale}} & \textbf{Temporality} & \textbf{\makecell{Spatial\\Grounding}} & \textbf{\makecell{Metric}} \\
\midrule
\midrule
Video-MME~\cite{fu2025video} & 900 & 2,700 & Close & Human & \xmark & \xmark & \xmark & \xmark & Accuracy \\
MMVU~\cite{zhao2025mmvu} & 1,529 & 3,000 & Open/Close & Human & \xmark & \cmark & \cmark & \xmark & Accuracy \\
LongVideoBench~\cite{wu2024longvideobench} & 3,763 & 6,678 & Close & Human & \xmark & \xmark & \cmark & \xmark & Accuracy \\
\midrule
EgoSchema~\cite{mangalamegoschema} & 250 hours+ & 5,000+ & Close & Human & \cmark & \xmark & \cmark & \xmark & Accuracy \\
EgoThink~\cite{Cheng_2024_CVPR} & 595 & 700 & Open & Human & \cmark & \xmark & \xmark & \xmark & LLM-Judge \\
EgoTempo~\cite{plizzari2024egptempo} & 365 & 500 & Open & Auto\&Human & \cmark & \xmark & \cmark & \xmark & LLM-Judge \\
MultiHop-EgoQA~\cite{chen2025grounded} & 360 & 1,080 & Open & Auto\&Human & \cmark & \xmark & \cmark & \xmark & Accuracy/LLM-Judge \\
EOC-Bench~\cite{yuan2025eocbench} & 656 & 3,277 & Open/Close & Human & \cmark & \xmark & \cmark & \cmark & Accuracy \\
EASG-Bench~\cite{Rodin_2025_ICCV} & 221 & 1,807 & Open & Auto & \cmark & \xmark & \cmark & \cmark & LLM-Judge \\
\midrule
\textbf{EgoCoT-Bench (Ours)} & \textbf{351} & \textbf{3,172} & \textbf{Open/Close} & \textbf{Auto\&Human} & \textbf{\cmark} & \textbf{\cmark} & \textbf{\cmark} & \textbf{\cmark} & \textbf{Accuracy/LLM-Judge} \\
\bottomrule
\end{tabular}%
}
\end{table*}
However, understanding dynamic object interactions in egocentric videos remains particularly challenging. Owing to the first-person viewpoint, manipulated objects are often only partially visible, intermittently leave and re-enter the field of view, and are frequently occluded by the wearer’s hands under rapid camera motion. The problem is further compounded by cluttered scenes and the presence of visually similar objects, which make the correct interaction target difficult to identify from instantaneous appearance alone~\cite{zhang2023learning, zhang2022boostmis, zhang2024revisiting}. More fundamentally, answering egocentric questions requires reasoning over temporally evolving evidence rather than relying solely on the current frame, including prior contact history, earlier object states, and the immediate context of an ongoing manipulation sequence~\cite{Sener_2022_CVPR,Wang_2023_ICCV,Di_2024_CVPR}.

Despite growing interest in egocentric understanding, existing benchmarks still suffer from \textbf{limited grounded rationale evaluation}. As summarized in Table~\ref{tab:benchmark_comparison}, general video benchmarks such as Video-MME~\cite{fu2025video}, MMVU~\cite{zhao2025mmvu}, and LongVideoBench~\cite{wu2024longvideobench} have substantially advanced video QA and temporal reasoning, but they are not designed for first-person interaction understanding and provide limited support for spatial grounding. Egocentric benchmarks such as EgoSchema~\cite{mangalamegoschema}, EgoThink~\cite{Cheng_2024_CVPR}, EgoTempo~\cite{plizzari2024egptempo}, and MultiHop-EgoQA~\cite{chen2025grounded} move evaluation closer to first-person settings, especially for temporal or open-ended reasoning, but they still provide limited support for explicit rationale supervision and fine-grained spatial grounding. More recent benchmarks such as EOC-Bench~\cite{yuan2025eocbench} and EASG-Bench~\cite{Rodin_2025_ICCV} further incorporate egocentric temporal and spatial evaluation, but they still offer limited support for jointly assessing rationale faithfulness, temporal sensitivity, and evidence-aware evaluation.

\begin{figure}[t]
    \centering

    % top label strip (no subfigure caption, no (a)(b))
    \includegraphics[width=\columnwidth]{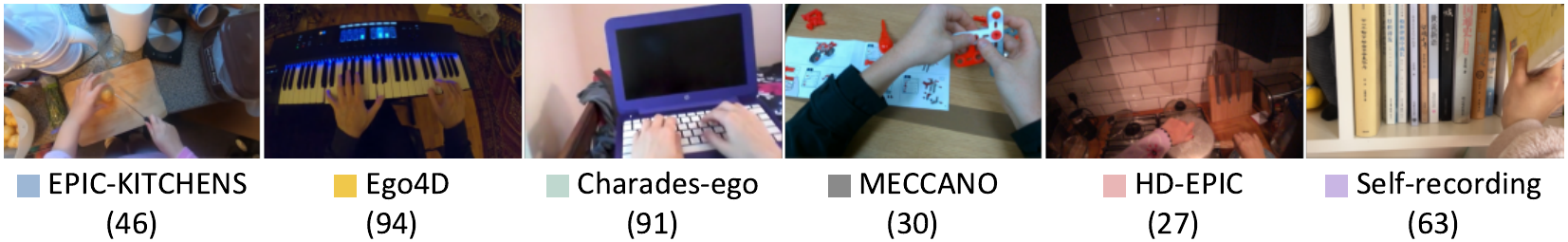}

    % bottom two pie charts
    \begin{subfigure}[t]{0.49\columnwidth}
        \centering
        \includegraphics[width=\linewidth]{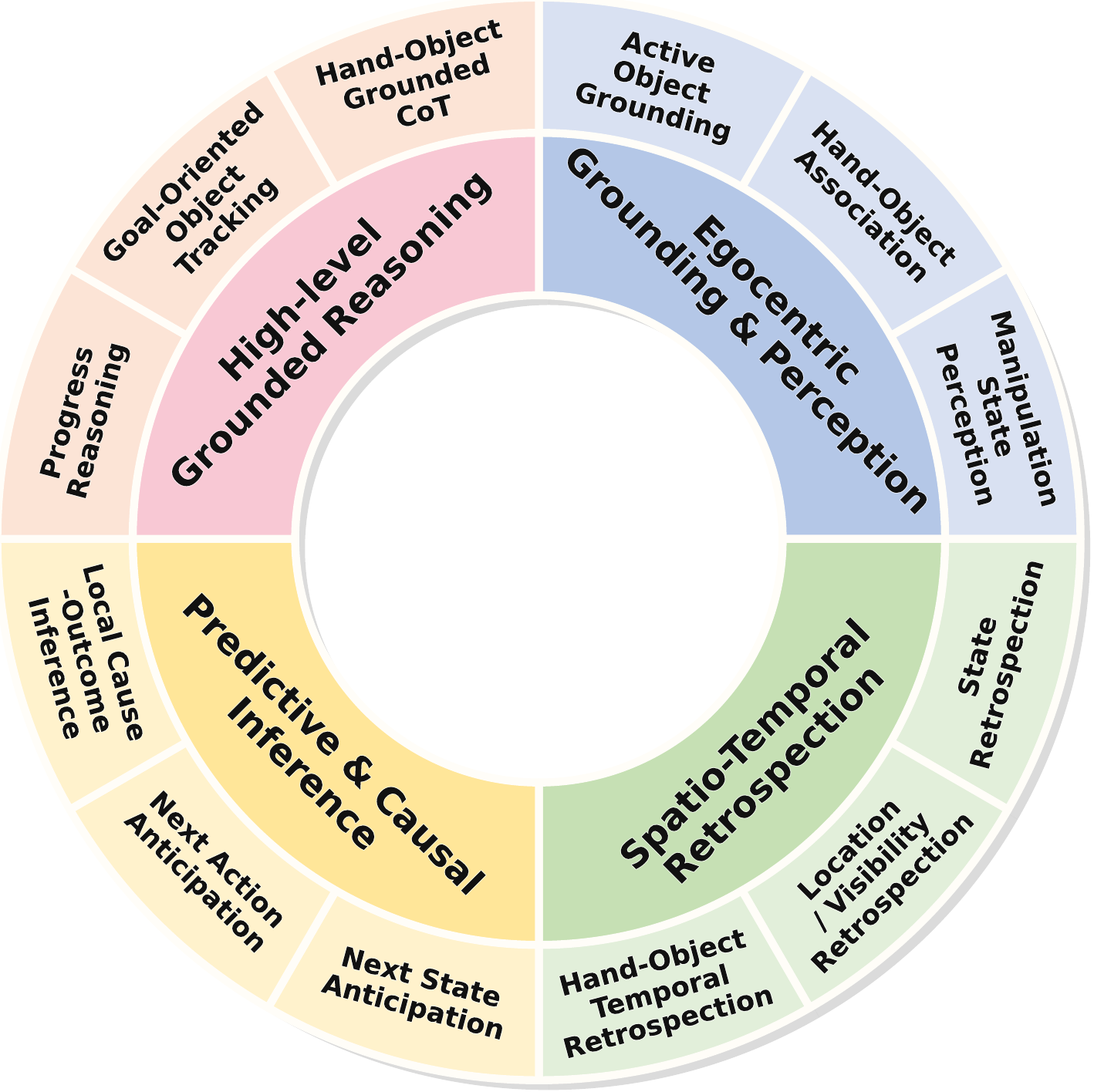}
        \caption{Dimensions of EgoCoT-Bench}
        \label{fig:benchmark_dimensions}
    \end{subfigure}
    \hfill
    \begin{subfigure}[t]{0.49\columnwidth}
        \centering
        \includegraphics[width=\linewidth]{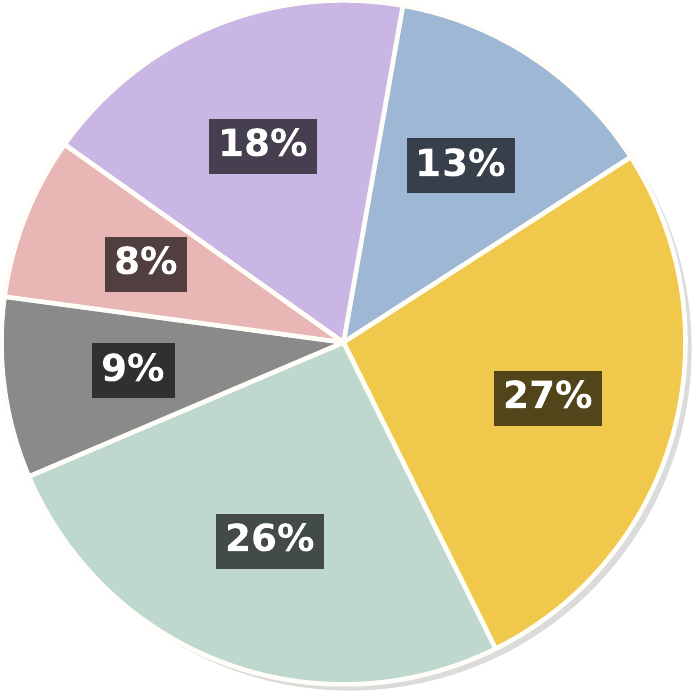}
        \caption{Video source distribution}
        \label{fig:benchmark_distribution}
    \end{subfigure}

    \caption{Overall statistics of EgoCoT-Bench. Top: representative video sources in the benchmark. (a) Dimensions of EgoCoT-Bench. (b) Distribution of EgoCoT-Bench samples.}
    \label{fig:benchmark_overall}
    \vspace{-0.8em}
\end{figure}
To address this gap, we introduce \textbf{EgoCoT-Bench}, a fine-grained egocentric benchmark for grounded and verifiable operation-centric reasoning with explicit step-by-step rationale annotations and spatio-temporal grounding. EgoCoT-Bench contains 3,172 QA pairs over \textbf{351} egocentric videos and is organized into four task groups with \textbf{12} fine-grained subtasks. These tasks cover egocentric grounding and perception, spatio-temporal retrospection, predictive and causal inference, and high-level grounded reasoning, targeting key capabilities required for first-person manipulation understanding beyond generic scene comprehension.

A central design goal of EgoCoT-Bench is to evaluate not only answer correctness but whether MLLMs reasoning are grounded in explicit first-person evidence. To this end, we construct the benchmark using a spatio-temporal scene graphs~(STSG)-guided generation framework. Candidate QA samples are first derived from structured egocentric interaction traces, and subsequently refined through human annotation to ensure semantic correctness, first-person relevance, and fine-grained reasoning quality. Each accepted sample is further augmented with structured evidence annotations—including timestamps, object identities, interaction relations, action history, and localized bounding boxes-enabling evaluation at both the answer and the evidence grounding level.

Using EgoCoT-Bench, we benchmark a range of representative MLLMs such as GPT~\cite{openai_gpt51,openai_gpt52}, Qwen~\cite{Qwen3-VL,qwen3.5} and LLaVA~\cite{li2024llava,LLaVA-OneVision-1.5} series, and observe that fine-grained egocentric reasoning remains highly challenging. While many models can produce correct answers, their underlying rationales are often temporally incomplete, weakly grounded, or inconsistent with the available object-level spatio-temporal evidence. This reveals a notable gap between answer correctness and reasoning faithfulness in current models, which may in turn limit performance gains and lead to error accumulation in more complex scenarios.
% These findings suggest that progress in egocentric video understanding should be measured not only by final answer accuracy, but also by whether the underlying reasoning is faithful to the spatio-temporal evidence presented in the video.
Our findings highlight the importance of moving beyond final answer accuracy, advocating instead for evaluation protocols that explicitly assess whether model reasoning is consistent with the underlying spatio-temporal evidence in egocentric video understanding.

In summary, our contributions are three-fold: \textbf{(1)} we introduce \textbf{EgoCoT-Bench}, a fine-grained egocentric benchmark for operation-centric reasoning, comprising 3,172 QA pairs over 351 videos across 12 subtasks; \textbf{(2)} we construct the benchmark via an STSG-guided generation and human refinement pipeline, with temporal and spatial evidence attached to each accepted samples for grounded first-person reasoning; and \textbf{(3)} we propose an evaluation protocol that jointly measures answer correctness, reasoning quality, and spurious correctness for a more faithful assessment.

% In summary, our contributions are three-fold:
% \begin{itemize}
%     \item We introduce \textbf{EgoCoT-Bench}, a fine-grained egocentric benchmark for operation-centric understanding, covering 3,172 QA pairs over \textbf{351} videos and 12 fine-grained subtasks.
%     \item We construct the benchmark through an STSG-guided generation and human refinement pipeline, and associate accepted samples with structured temporal and spatial evidence for fine-grained first-person reasoning.
%     \item We introduce an evaluation protocol that jointly measures answer correctness, reasoning quality, and spurious correctness, enabling a more faithful assessment of grounded and verifiable reasoning in egocentric video understanding.
% \end{itemize}

\section{Related Work}

\subsection{Egocentric Video Understanding}

Egocentric video understanding has received increasing attention in recent years, driven by its importance for embodied AI, assistive systems, and first-person human activity analysis~\cite{Damen_2018_ECCV, Grauman_2022_CVPR, Li_2018_ECCV, sigurdsson2018actor,li2018eye}. Large-scale datasets such as Ego4D and EPIC-KITCHENS have advanced research on egocentric perception, activity recognition, narration, forecasting, and long-form video understanding~\cite{ragusa2021meccano, ragusa2022meccano, Sener_2022_CVPR, Wang_2023_ICCV, VISOR2022,Damen2022RESCALING,Grauman_2024_CVPR}. Beyond these foundational resources, more recent benchmarks have extended evaluation toward egocentric question answering, scene-text understanding, object-centric cognition, and cross-view reasoning~\cite{Zhou_2025_CVPR,yuan2025eocbench,he2025egoexobench,Majumdar_2024_CVPR,jia2022egotaskqa,chen2024egoplanbenchbenchmarkingmultimodallarge,yuan2025videorefer,yuan2026lmms,yuan2025pixelrefer}. These benchmarks have played an important role in promoting first-person video understanding, but many of them primarily emphasize broad scene understanding, long-context comprehension, or object-centric reasoning at a relatively coarse level~\cite{Di_2024_CVPR,mangalamegoschema}. As a result, they are less suited for systematically evaluating fine-grained operation-centric reasoning in dynamic first-person scenarios, such as identifying active manipulation targets, tracking short-horizon state changes, or recovering temporally localized hand-object interaction evidence.

\subsection{Video Reasoning Benchmarks}

A large body of work has also studied reasoning-oriented evaluation for general video understanding~\cite{li2023mvbench,liu2024tempcompass,wu2024longvideobench}. Existing benchmarks have explored temporal perception, event ordering, motion understanding, long-context reasoning, and multi-step video question answering~\cite{zhou2024mlvu,liu2024tempcompass,wu2024longvideobench,cheng2025vstarbenchmarkingvideollmsvideo,wu2021star_situated_reasoning}. These resources have significantly improved the diagnosis of multimodal reasoning ability in video-based settings, especially for temporal comprehension and general event-level inference~\cite{Xiao_2021_CVPR,chen2024rextime}. However, compared with egocentric manipulation scenarios, generic video reasoning benchmarks are typically less sensitive to the distinctive challenges of first-person interaction, where reasoning often depends on local object contact, operator viewpoint, immediate action history, and subtle state transitions. Consequently, they provide only limited support for evaluating whether a model can reason over object-centered manipulation processes in a temporally and spatially grounded manner.

% \subsection{Grounded and Verifiable Reasoning}

% Recent progress in multimodal reasoning has led to growing interest in chain-of-thought supervision, rationale generation, and explanation-oriented evaluation for vision-language models~\cite{cot,mm-cot,video-cot}. At the same time, structured representations such as scene graphs and spatio-temporal scene graphs have been widely studied as a way to organize objects, relations, and temporal dynamics in videos~\cite{scenegraph,stsg}. These directions suggest that reasoning quality should ideally be evaluated not only by the final answer, but also by whether the underlying rationale is faithful to explicit visual evidence. Nevertheless, existing video and egocentric benchmarks rarely provide a benchmark setting that jointly supports first-person fine-grained reasoning, temporally sensitive interaction understanding, and evidence-aware evaluation with localized object grounding. In contrast, our benchmark is designed to fill this gap by focusing on operation-centric egocentric QA with structured evidence support, enabling evaluation at both the answer level and the evidence level.

\section{EgoCoT-Bench}

\subsection{Overview}

We introduce \textbf{EgoCoT-Bench}, a fine-grained benchmark for egocentric video understanding that focuses on operation-centric reasoning in dynamic first-person environments. EgoCoT-Bench contains 3,172 QA pairs collected from \textbf{351} egocentric video clips. It is organized into four task groups, covering a total of 12 fine-grained subtasks. These tasks are designed to systematically assess egocentric grounding and perception, temporal retrospection, predictive and causal inference, and high-level grounded reasoning. Together, they target a core challenge of first-person understanding: whether a model can reason about dynamic object-centered interactions in a temporally and spatially grounded manner.
\begin{table*}[t]
\caption{Main results on EgoCoT-Bench. Results are reported as accuracy (\%).}
\label{tab:main_results}
\centering
% 可以保留 \small 配合 resizebox 获得最佳基础比例
\small
\setlength{\tabcolsep}{3.2pt}
% ================= 核心修改：强制缩放至页面宽度 =================
\resizebox{0.98\textwidth}{!}{%
\renewcommand{\arraystretch}{1.00}
\begin{tabular}{l|c|cccc|cccc|cccc|cccc}
\toprule
\multirow{2}{*}{\textbf{Method}} & \multirow{2}{*}{\textbf{Mean}} 
& \multicolumn{4}{c|}{\textbf{\makecell{Egocentric\\ Grounding \& Perception}}} 
& \multicolumn{4}{c|}{\textbf{\makecell{Spatio-Temporal\\Retrospection}}}
& \multicolumn{4}{c|}{\textbf{\makecell{Predictive \& Causal\\Inference}}}
& \multicolumn{4}{c}{\textbf{\makecell{High-level Grounded\\Reasoning}}}\\
% ================= 核心修改：修正并启用了断线 =================
\cmidrule(lr){3-6} \cmidrule(lr){7-10} \cmidrule(lr){11-14} \cmidrule(lr){15-18}
& 
& \textbf{AOG} & \textbf{HOA} & \textbf{MSP} & \textbf{Mean}
& \textbf{SR} & \textbf{LVR} & \textbf{HOTR} & \textbf{Mean}
& \textbf{NSA} & \textbf{NAA} & \textbf{LCOI} & \textbf{Mean}
& \textbf{PR} & \textbf{HGC} & \textbf{GOT} & \textbf{Mean}\\
\midrule
\midrule
\rowcolor{humanbg}
Human & 95.93 & 96.18 & 93.86 & 97.18 & 96.11 & 93.96 & 95.36 & 96.88 & 94.96 & 98.47 & 94.44 & 95.00 & 96.32 & 98.71 & 97.90 & 91.98 & 96.34 \\
\midrule
\rowcolor{gray!10}
\multicolumn{18}{c}{\textit{Proprietary Multimodal Foundation Models}} \\
\midrule
GPT-5.1~\cite{openai_gpt51} & 66.71 & 64.20 & \best{63.16} & 77.00 & 67.69 & 66.04 & 49.40 & 56.25 & 55.23 & \best{86.99} & 67.17 & 70.56 & 76.63 & 68.24 & 79.83 & \best{45.28} & 65.15 \\
GPT-5.2~\cite{openai_gpt52} & 67.91 & 64.92 & \second{62.28} & 72.30 & 66.62 & 67.55 & 59.88 & 59.38 & 62.42 & \second{84.69} & 64.14 & 72.22 & 75.68 & 65.24 & \best{84.03} & \second{42.92} & 64.86 \\
Qwen3-VL-Plus~\cite{Qwen3-VL} & 67.12 & \second{69.21} & 61.40 & 77.00 & 70.24 & 69.70 & 54.66 & 56.25 & 59.75 & \second{84.69} & 66.16 & 71.94 & 76.00 & 68.53 & 77.54 & 32.70 & 60.53 \\
Qwen3.5-Plus~\cite{qwen3.5} & \second{70.68} & 68.26 & \second{62.28} & 85.92 & \best{72.39} & 67.55 & 60.69 & 53.12 & 62.67 & 85.20 & \best{70.71} & \best{74.72} & \best{78.21} & \best{81.12} & 82.77 & 35.85 & \second{67.64} \\
\midrule
\rowcolor{gray!10}
\multicolumn{18}{c}{\textit{Open-Source Multimodal Foundation Models}} \\
\midrule
InternVL3.5-1B~\cite{wang2025internvl3} & 53.91 & 41.77 & 55.26 & 69.95 & 51.88& 44.91 & 51.21 & \best{75.00} & 50.06 & 64.29 & 55.56 & 56.94 & 59.68 & 55.36 & 67.65 & 32.55 & 52.56 \\
InternVL3.5-2B~\cite{wang2025internvl3} & 61.79 & 50.60 & \best{63.16} & 79.81 & 60.86 & 58.11 & \second{66.94} & \second{71.88} & 64.18 & 77.30 & 59.09 & 58.33 & 66.32 & 60.94 & 71.01 & 26.42 & 53.73 \\
InternVL3.5-4B~\cite{wang2025internvl3} & 61.95 & 58.71 & 59.65 & 73.24 & 63.00 & 48.30 & 54.44 & 62.50 & 52.71 & 81.63 & 59.60 & 63.06 & 70.00 & 64.81 & 75.63 & 38.21 & 60.32 \\
LLaVA-OneVision-1.5-4B~\cite{LLaVA-OneVision-1.5} & 60.78 & 55.74 & 54.39 & 72.30 & 60.27 & 61.51 & 55.04 & 40.62 & 56.62 & 81.38 & 61.11 & 61.67 & 69.68 & 63.95 & 73.11 & 21.23 & 53.88 \\
LLaVA-NeXT-Video-7B~\cite{li2024llava} & 44.26 & 35.08 & 55.26 & 46.95 & 41.55 & 46.42 & 40.93 & 56.25 & 43.38 & 62.24 & 48.99 & 48.61 & 54.32 & 33.91 & 49.58 & 17.45 & 34.26 \\
InternVL3.5-8B~\cite{wang2025internvl3} & 64.06 & 56.32 & 61.40 & 70.89 & 61.26 & 54.34 & \best{67.74} & 68.75 & 63.30 & 80.36 & 66.16 & 67.22 & 72.42 & 66.95 & 73.11 & 25.94 & 56.37 \\
LLaVA-OneVision-1.5-8B~\cite{LLaVA-OneVision-1.5} & 60.81 & 53.94 & 58.77 & 74.65 & 60.59 & 57.36 & 51.61 & 40.62 & 53.09 & 82.14 & 69.19 & 61.94 & 71.79 & 68.67 & 71.01 & 21.23 & 54.76 \\
Qwen3-VL-8B~\cite{Qwen3-VL} & 65.42 & \best{69.54} & 59.29 & 81.60 & 71.43 & 64.02 & 60.69 & 59.38 & 61.75 & 83.03 & 63.13 & 64.72 & 71.91 & 59.91 & 78.15 & 25.00 & 55.43 \\
InternVL3.5-14B~\cite{wang2025internvl3} & 64.09 & 56.32 & 56.14 & 75.12 & 61.66 & 56.98 & 63.71 & \best{75.00} & 61.92 & 78.32 & 65.66 & 70.00 & 72.53 & 61.37 & 72.27 & 36.79 & 57.54 \\
Qwen3.5-27B~\cite{qwen3.5} & \best{71.28} & 68.26 & 61.40 & 84.51 & \second{71.85} & \best{72.83} & \best{67.74} & 59.38 & \best{69.10} & 84.65 & 68.69 & 73.33 & \second{77.03} & 77.68 & 80.67 & 34.43 & 65.30 \\
Qwen3-VL-30B-A3B~\cite{Qwen3-VL} & 64.63 & 62.44 & 61.06 & 82.16 & 67.88 & 65.15 & 65.86 & 62.50 & \second{65.49} & 81.89 & 59.09 & 66.94 & 71.47 & 66.52 & 73.11 & 9.05 & 51.10 \\
Qwen3-VL-32B~\cite{Qwen3-VL} & 67.09 & 67.78 & \second{62.28} & 79.81 & 70.38 & 64.53 & 55.04 & 68.75& 58.76 & 84.18 & \second{70.20} & 71.67 & 76.53 & 69.10 & 79.83 & 27.83 & 60.03 \\
Qwen3.5-122B-A10B~\cite{qwen3.5} & 69.96 & 68.26 & 61.40 & \second{86.38} & \best{72.39} & 70.72 & 62.70 & 56.25 & 65.11 & 81.63 & \best{70.71} & \second{73.61} & 76.32 & \second{79.83} & 79.41 & 30.19 & 64.28 \\
Qwen3-VL-235B-A22B~\cite{Qwen3-VL} & 65.86 & 67.54 & 57.02 & 77.00 & 68.63 & \second{71.32} & 52.82 & 56.25 & 59.14 & 85.97 & 68.18 & 62.50 & 73.37 & 70.82 & 78.99 & 27.36 & 60.18 \\
Qwen3.5-397B-A17B~\cite{qwen3.5}o & 70.11 & 68.26 & 58.77 & \best{87.79} & \best{72.39} & 69.81 & 59.07 & 56.25 & 62.55 & 84.95 & 68.18 & 70.83 & 76.11 & 78.97 & 83.61 & 38.68 & \best{68.08} \\
\bottomrule
\end{tabular}%
} % 结束 resizebox
\end{table*}

\subsection{Benchmark Construction}

\subsubsection{Video Collection}

To ensure both diversity and task relevance, the video collection of EgoCoT-Bench is curated from a wide range of egocentric sources. Specifically, we integrate public first-person datasets including \textbf{Ego4D}~\cite{Grauman_2022_CVPR}, \textbf{EPIC-KITCHENS}~\cite{Damen_2018_ECCV}, \textbf{MECCANO}~\cite{ragusa2022meccano, ragusa2021meccano}, \textbf{Charades-Ego}~\cite{sigurdsson2018actor}, and \textbf{HD-EPIC}~\cite{perrett2025hdepic}, together with a supplementary set of \textbf{self-recorded videos}. These sources provide complementary coverage of interaction scenarios, ranging from daily object use and kitchen activities to more structured manipulation and assembly processes, thereby supporting a comprehensive evaluation of egocentric reasoning tasks.

% To improve suitability for benchmark construction, the raw videos are further processed into shorter clips centered around localized manipulation events. This design allows us to focus the benchmark on temporally coherent first-person interactions, where the key evidence often lies in short but critical windows of contact, movement, and state transition. As a result, [BenchmarkName] captures a wide range of egocentric operation scenarios while maintaining the temporal granularity required for fine-grained reasoning.

\subsubsection{Task Taxonomy}

To systematically characterize first-person operation-centric understanding, we organize EgoCoT-Bench into four task groups with twelve fine-grained subtasks.
\paragraph{(i) Egocentric Grounding \& Perception.}
This group evaluates current interaction grounding in first-person videos: \textbf{Active Object Grounding (AOG)} identifies the object currently attended to, touched, or manipulated by the operator; \textbf{Hand-Object Association (HOA)} determines which hand is interacting with which object; and \textbf{Manipulation State Perception (MSP)} recognizes the current manipulation-related state of the object.

\paragraph{(ii) Spatio-Temporal Retrospection.}
This group measures whether a model can recover object-centric evidence from preceding moments: \textbf{State Retrospection (SR)} recalls an object’s earlier state; \textbf{Location / Visibility Retrospection (LVR)} recovers its previous location or status; and \textbf{Hand-Object Temporal Retrospection (HOTR)} infers the temporal order of hand-object interactions.

\paragraph{(iii) Predictive \& Causal Inference.}
This group evaluates short-horizon anticipation and local causal reasoning grounded in the current manipulation context: \textbf{Next State Anticipation (NSA)} predicts an object’s most likely next state; \textbf{Next Action Anticipation (NAA)} predicts the operator’s most likely next action; and \textbf{Local Cause-Outcome Inference (LCOI)} identifies the recent action directly responsible for the observed outcome or state change.

\paragraph{(iv) High-level Grounded Reasoning.}
This group focuses on compositional reasoning over progress, evidence chains, and goal-oriented tracking: \textbf{Progress Reasoning (PR)} infers the current operation step or whether a step has been completed; \textbf{Hand-Object Grounded CoT (HGC)} generates interpretable reasoning chains that combine hand-object cues, temporal evidence, and visual grounding; and \textbf{Goal-Oriented Object Tracking (GOT)} tracks an object over time according to its functional role in the ongoing manipulation goal.
\subsubsection{Construction Pipeline}

\paragraph{STSG-Guided Candidate Generation}
To ensure the quality and verifiability of EgoCoT-Bench, we build the benchmark through a structured human-in-the-loop pipeline in which candidate generation is grounded in verified spatio-temporal scene graph (STSG) rather than free-form video description as illustrated in Figure~\ref{fig:overview}. In the first stage, each video clip is converted into an ego-adapted STSG, which serves as an intermediate representation for candidate construction. Before candidate generation, the STSG is manually inspected and refined to correct unreliable object identities, temporally inconsistent interaction links, ambiguous state transitions, and misaligned spatial grounding. The STSG organizes object instances, operator body parts, action traces, interaction relations, temporal states, and bounding boxes across time. 
\begin{table*}[t]
\caption{Reasoning Score (R) and Spurious Correct Rate (SCR) evaluation on EgoCoT-Bench. Results are reported with R on a strict 0-5 scale and SCR in percentage (\%). Higher R is better, while higher SCR indicates worse answer-reasoning consistency.}
\label{tab:reasoning_score}
\centering
\small
\setlength{\tabcolsep}{3.2pt}
\resizebox{0.93\textwidth}{!}{%
\renewcommand{\arraystretch}{1.00}
\begin{tabular}{l|cc|C{1.6cm}C{1.6cm}|C{1.3cm}C{1.3cm}|C{1.4cm}C{1.4cm}|C{1.4cm}C{1.4cm}}
\toprule
\multirow{2}{*}{\textbf{Method}} 
& \multicolumn{2}{c|}{\textbf{Mean}} 
& \multicolumn{2}{c|}{\textbf{\makecell{Egocentric\\ Grounding \& Perception}}} 
& \multicolumn{2}{c|}{\textbf{\makecell{Spatio-Temporal\\Retrospection}}} 
& \multicolumn{2}{c|}{\textbf{\makecell{Predictive \& Causal\\Inference}}} 
& \multicolumn{2}{c}{\textbf{\makecell{High-level Grounded\\Reasoning}}}\\
\cmidrule(lr){2-3} \cmidrule(lr){4-5} \cmidrule(lr){6-7} \cmidrule(lr){8-9} \cmidrule(lr){10-11}
& \textbf{R $\uparrow$} & \textbf{SCR $\downarrow$} 
& \textbf{R $\uparrow$} & \textbf{SCR $\downarrow$} 
& \textbf{R $\uparrow$} & \textbf{SCR $\downarrow$} 
& \textbf{R $\uparrow$} & \textbf{SCR $\downarrow$} 
& \textbf{R $\uparrow$} & \textbf{SCR $\downarrow$}\\
\midrule
\midrule
\rowcolor{gray!10}
\multicolumn{11}{c}{\textit{Proprietary Multimodal Foundation Models}} \\
\midrule
GPT-5.1& 2.77 & \second{4.91} & 2.65 & \second{5.35} & 2.16 & 5.25 & 3.43 & \best{1.79} & 2.67 & 9.21 \\
GPT-5.2& 2.85 & \best{4.27} & 2.39 & \best{3.23} & \best{2.73} & \best{4.02} & 3.40 & \second{2.36} & 2.76 & \second{8.80} \\
Qwen3-VL-Plus& \best{3.08} & 7.84 & \best{3.04} & 7.44 & \second{2.61} & 5.29 & \best{3.64} & 6.37 & \second{2.87} & 13.87 \\
Qwen3.5-Plus& 2.92 & 9.10 & \second{2.88} & 7.41 & 2.31 & 10.87 & 3.50 & 7.67 & \second{2.87} & 11.47 \\
\midrule
\rowcolor{gray!10}
\multicolumn{11}{c}{\textit{Open-Source Multimodal Foundation Models}} \\
\midrule
InternVL3.5-1B& 2.21 & 13.33 & 2.11 & 8.53 & 1.77 & 17.63 & 2.70 & 8.99 & 2.14 & 20.61 \\
InternVL3.5-2B& 2.53 & 7.86 & 2.43 & 9.69 & 2.29 & 7.86 & 2.98 & 2.70 & 2.29 & 14.44 \\
InternVL3.5-4B& 2.45 & 9.57 & 2.39 & 8.72 & 1.86 & 9.57 & 3.09 & 4.96 & 2.31 & 17.96 \\
LLaVA-OneVision-1.5-4B& 2.50 & 9.57 & 2.32 & 10.47 & 2.03 & 14.92 & 3.17 & 3.77 & 2.32 & 13.59 \\
LLaVA-NeXT-Video-7B& 1.85 & 22.93 & 1.57 & 25.16 & 1.51 & 35.46 & 2.59 & 8.53 & 1.53 & 33.33 \\
InternVL3.5-8B& 2.56 & 5.61 & 2.39 & 5.47 & 2.25 & 4.98 & 3.15 & 3.92 & 2.30 & 9.61 \\
LLaVA-OneVision-1.5-8B& 2.21 & 24.73 & 2.08 & 28.31 & 1.76 & 27.31 & 2.77 & 21.11 & 2.08 & 24.06 \\
Qwen3-VL-8B& 2.73 & 10.07 & 2.74 & 9.25 & 2.29 & 14.40 & 3.27 & 6.46 & 2.47 & 12.17 \\
InternVL3.5-14B& 2.60 & 5.36 & 2.50 & 5.43 & 2.27 & \second{4.48} & 3.17 & 2.46 & 2.30 & 11.45 \\
Qwen3.5-27B& 2.96 & 7.25 & 2.87 & 8.39 & 2.49 & 8.94 & 3.56 & 3.15 & 2.78 & 10.54 \\
Qwen3-VL-30B-A3B& 2.79 & 7.25 & 2.73 & 8.91 & 2.43 & 10.42 & 3.36 & 5.89 & 2.46 & \best{7.76} \\
Qwen3-VL-32B& \second{2.96} & 7.99 & \second{2.88} & 9.90 & 2.40 & 7.51 & \second{3.63} & 5.36 & 2.77 & 10.73 \\
Qwen3.5-122B-A10B& 2.94 & 9.73 & 2.83 & 11.29 & 2.43 & 11.26 & 3.48 & 7.45 & \best{2.88} & 9.79 \\
Qwen3-VL-235B-A22B& 2.78 & 11.01 & 2.70 & 11.32 & 2.32 & 9.38 & 3.42 & 8.90 & 2.53 & 16.05 \\
Qwen3.5-397B-A17B& 2.87 & 10.93 & 2.86 & 9.81 & 2.29 & 12.70 & 3.37 & 9.82 & \second{2.87} & 12.04 \\
\bottomrule
\end{tabular}%
}
\end{table*}
Based on the refined STSG, we derive candidate samples by traversing task-specific evidence paths that connect each target answer to concrete first-person interaction cues. The LLM is then used to render these verified structural facts into natural-language questions, answer options, and rationales, rather than to invent the underlying evidence. For every candidate sample, we preserve the associated structural metadata, including timestamps, object identities, action history, interaction relations, and bounding boxes, so that the sample remains traceable and can be explicitly checked during downstream evidence-aware evaluation.

\paragraph{Human Refinement and Quality Control}
All generated candidates are then subjected to careful manual screening under a multi-round review protocol. First, four human annotators independently perform an initial screening pass to remove obviously invalid or weak candidates, such as those with ambiguous targets, weak first-person relevance, inconsistent reasoning, or low-quality distractors. Next, the retained candidates are cross-checked by different reviewers, who verify the consistency among the question, answer, and reasoning. Finally, a lead reviewer performs the last-round inspection and adjudication, resolving disagreements, rejecting low-confidence cases, and confirming the final accepted version. We keep a sample only when its question, answer, rationale, and supporting evidence are mutually consistent and clearly grounded in the video. Through this process, EgoCoT-Bench retains only samples that satisfy semantic correctness, egocentric relevance, and evidence consistency, yielding a human-refined benchmark with structured temporal and spatial evidence support.

\subsubsection{Evaluation Metrics}
\label{sec:eval_metrics}

To provide a comprehensive assessment of multimodal large language models (MLLMs) in egocentric environments, EgoCoT-Bench evaluates not only the final answer correctness but also the quality of the reasoning process and the consistency between them. Conventional benchmarks often rely solely on answer accuracy, which may overestimate model capability when the correct answer is obtained without sound reasoning. To address this issue, we adopt a three-metric evaluation protocol consisting of Answer Accuracy (Acc), Reasoning Score (R), and Spurious Correct Rate (SCR).

\paragraph{\textbf{Answer Accuracy (Acc)}}
All tasks in EgoCoT-Bench are formulated as four-way multiple-choice questions. We adopt a strict exact-match criterion to evaluate the final prediction.
% Given a dataset of $N$ samples, let $y_i$ denote the ground-truth answer and $\hat{y}_i$ denote the model prediction. The Answer Accuracy is defined as:
% \begin{equation}
%     \text{Acc} = \frac{1}{N} \sum_{i=1}^{N} \mathbb{I}(\hat{y}_i = y_i),
% \end{equation}
% where $\mathbb{I}(\cdot)$ is the indicator function.

\paragraph{\textbf{Reasoning Score (R)}}
In egocentric video understanding, predictions may be unsupported by grounded and coherent reasoning. We therefore evaluate the model's reasoning quality by scoring its generated reasoning steps against the annotated reference reasoning by employing a strong LLM (Qwen-Max) as a judge to assess each prediction from the perspectives of logical coherence, factual consistency, and alignment with the visual evidence on a 0-5 scale.
% The detailed prompt and scoring rubric are provided in the appendix.
% The detailed scoring rubric is provided in the appendix. The overall Reasoning Score is computed as:
% \begin{equation}
%     \text{R} = \frac{1}{N} \sum_{i=1}^{N} S_{\text{judge}}(\hat{c}_i, c_i),
% \end{equation}
% where $\hat{c}_i$ and $c_i$ denote the predicted and reference reasoning for sample $i$, respectively, and $S_{\text{judge}}(\hat{c}_i, c_i)$ denotes the judge-assigned reasoning score.

\paragraph{\textbf{Spurious Correct Rate (SCR)}}
% Although Acc reflects answer correctness and R captures reasoning quality, neither metric explicitly measures how often a model arrives at the correct answer with weak reasoning. 
To quantify how often a model arrives at the correct answer with weak reasoning, we introduce Spurious Correct Rate (SCR), which measures the proportion of answer-correct cases whose reasoning remains weak. Specifically, a prediction is considered \emph{spurious correct} if it satisfies both $\hat{y}_i = y_i$ and $S_{\text{judge}}(\hat{c}_i, c_i) \leq 2$. The SCR is defined as:
\begin{equation}
    \text{SCR} =
    \frac{
        \sum_{i=1}^{N}
        \mathbb{I}(\hat{y}_i = y_i)\,
        \mathbb{I}\!\left(S_{\text{judge}}(\hat{c}_i, c_i) \leq 2\right)
    }{
        \sum_{i=1}^{N} \mathbb{I}(\hat{y}_i = y_i)
    }.
\end{equation}
SCR is reported as a percentage where a higher value indicates worse answer-reasoning consistency.

% \subsection{Benchmark Statistics}

% [BenchmarkName] comprises \textbf{3,172} QA pairs in total, systematically evaluating MLLMs across \textbf{12} fine-grained subtasks under four task groups. Among them, \textbf{746} questions belong to Egocentric Grounding \& Perception, \textbf{793} to Spatio-Temporal Retrospection, \textbf{950} to Predictive \& Causal Inference, and \textbf{683} to High-level Grounded Reasoning. This distribution reflects our goal of covering both immediate first-person interaction understanding and more compositional reasoning over temporally evolving manipulation processes.

% In addition to answer annotations, each accepted sample is associated with structured evidence derived from the underlying ego-adapted STSG. Specifically, the benchmark incorporates temporally aligned object states, interaction relations, action traces, and localized bounding boxes, enabling evidence-aware evaluation beyond answer correctness alone. The benchmark contains \textbf{[O]} distinct object instances and \textbf{[C]} object categories, spanning a wide range of egocentric manipulation scenarios. Furthermore, \textbf{[T\%]} of the samples involve temporally sensitive reasoning, while \textbf{[S\%]} can be linked to explicit spatial grounding cues such as object-level localization or bounding-box-aligned evidence.

% Overall, these statistics show that [BenchmarkName] is not only a fine-grained egocentric QA benchmark, but also a structured testbed for studying temporally and spatially grounded reasoning in first-person videos.

\begin{figure*}[t]
    \centering
    \includegraphics[width=0.93\textwidth]{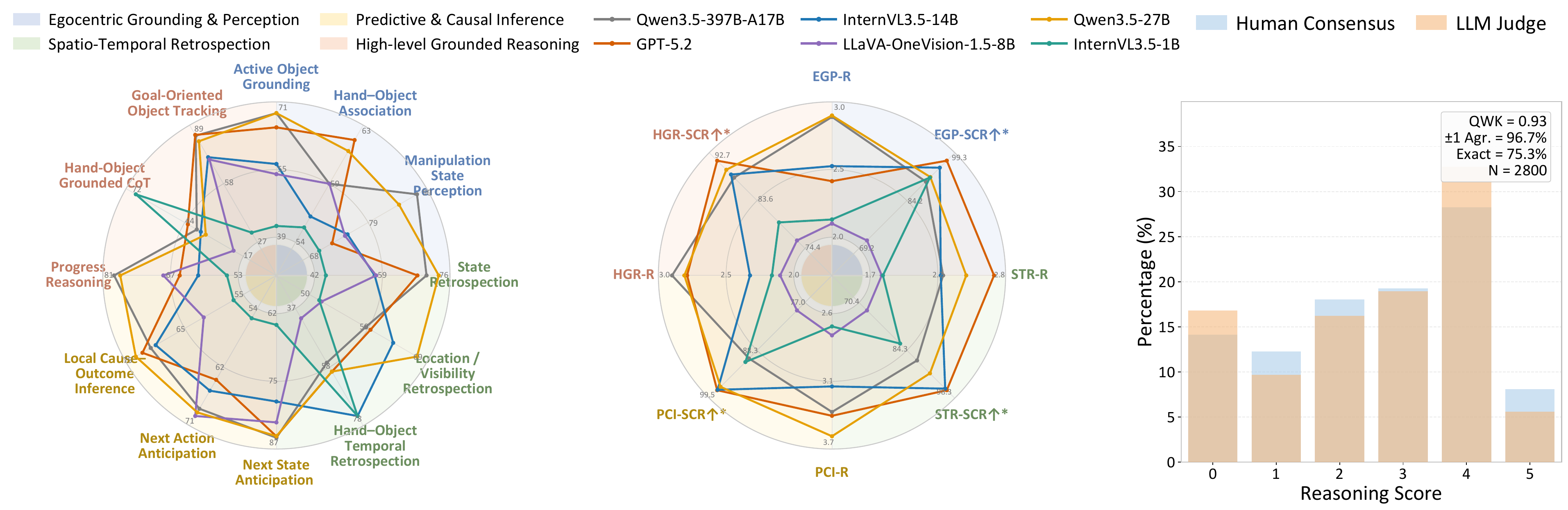}
    \caption{\textbf{Fine-grained radar analysis on EgoCoT-Bench.}
    Left: answer accuracy (\%) across 12 subtasks. Middle: reasoning quality across four task groups using Reasoning Score (R) and inverted Spurious Correct Rate (SCR↑*, i.e., 100-SCR), where larger radii indicate better performance. Right: comparison between human ratings and LLM-judge scores on a randomly sampled subset of model responses.}
    \label{fig:radar_analysis}
    \vspace{-0.6em}
\end{figure*}
\section{Experiment}
\subsection{Models and Human Evaluation.}
We evaluate a broad set of multimodal large language models (MLLMs) on \textbf{EgoCoT-Bench}, including 4 proprietary MLLMs and 15 open-source MLLMs spanning different parameter scales and architectural families. Among proprietary models, we evaluate GPT-5.1~\cite{openai_gpt51}, GPT-5.2~\cite{openai_gpt52}, Qwen3-VL-Plus~\cite{Qwen3-VL} and Qwen3.5-Plus~\cite{qwen3.5}. For open-source models, we test InternVL3.5~\cite{wang2025internvl3}, LLaVA-OneVision-1.5~\cite{LLaVA-OneVision-1.5}, LLaVA-NeXT-Video~\cite{li2024llava}, Qwen3-VL~\cite{Qwen3-VL} and Qwen3.5~\cite{qwen3.5}. In addition to model evaluation, we also measure human performance on EgoCoT-Bench with three volunteers.

\subsection{Main Results Analysis.}
The overall results in Table~\ref{tab:main_results} show that EgoCoT-Bench remains highly \textbf{challenging for current MLLMs}. The best overall accuracy is achieved by Qwen3.5-27B with Benchmark Models and Human Evaluation, followed by Qwen3.5-Plus with 70.68\% and Qwen3.5-397B-A17B with 70.11\%. However, even the strongest model still trails human performance (95.93\%) by a large margin, indicating that fine-grained egocentric reasoning is far from solved. This gap is consistently observed across all four task groups.

From a group-level perspective, Predictive \& Causal Inference is comparatively more tractable than the other dimensions, where the best group accuracy reaches 78.21\% by Qwen3.5-Plus. In contrast, Spatio-Temporal Retrospection and High-level Grounded Reasoning remain notably harder, with the best group results being only 69.10\% and 68.08\%, respectively. This pattern suggests that \textbf{current MLLMs are relatively better at short-horizon anticipation and local cause-outcome reasoning than at recovering earlier interaction evidence or performing compositional object-centered reasoning over longer temporal context.}

\paragraph{Fine-grained Task Analysis.}
Fig.~\ref{fig:radar_analysis} further reveals a highly uneven capability profile across the 12 subtasks. Among individual subtasks, models are relatively strong on Manipulation State Perception (MSP), Next State Anticipation (NSA), and Hand-Object Grounded CoT (HGC), where the best accuracies reach 87.79\%, 86.99\%, and 84.03\%, respectively. These results suggest that \textbf{current MLLMs can often capture immediate object state cues and some short-range action consequences when the visual evidence is sufficiently explicit.}

By contrast, Goal-Oriented Object Tracking (GOT) is by far the most difficult subtask. The best model achieves only 45.28\%, which is lower than the human score of 91.98\%. This large gap indicates that tracking an object according to its functional role in an evolving manipulation process is still beyond the capability of current systems. In addition, tasks such as Hand-Object Association (HOA) and Location / Visibility Retrospection (LVR) also remain challenging, suggesting \textbf{persistent weaknesses in local interaction grounding and in recalling object-centric evidence from earlier moments.}

\paragraph{Reasoning Quality and Answer-Reasoning Consistency.}
Table~\ref{tab:reasoning_score} shows that answer correctness and reasoning quality do not fully align. Although Qwen3.5-27B achieves the best overall accuracy, the highest mean reasoning score is obtained by Qwen3-VL-Plus with 3.08/5. Meanwhile, GPT-5.2 yields the lowest SCR at only 4.27\%, indicating the strongest consistency between correct answers and acceptable reasoning among the evaluated models. These results confirm that \textbf{a model may obtain the right answer while still relying on weak, incomplete, or weakly grounded rationales.} The right panel of Fig.~\ref{fig:radar_analysis} further shows that the LLM-judge is well aligned with human evaluation, as evidenced by a high quadratic weighted kappa (QWK = 0.93), 96.7\% $\pm1$ agreement, and 75.3\% exact agreement on 2,800 randomly selected responses.

This inconsistency becomes more evident in the group-wise reasoning analysis. On Predictive \& Causal Inference, several models achieve relatively high reasoning scores together with low SCR, suggesting that short-horizon causal judgments are easier to verbalize coherently. In contrast, High-level Grounded Reasoning exhibits substantially worse SCR for many models, despite moderate answer accuracy. This suggests that \textbf{models can sometimes guess the correct option, yet fail to provide reasoning that is faithfully aligned with the relevant hand-object interactions, temporal evidence, or functional object roles.}

Overall, these results highlight the importance of evaluating egocentric reasoning beyond answer accuracy alone. EgoCoT-Bench exposes a non-trivial amount of \emph{spurious correctness}, where answer-level success can mask insufficiently grounded reasoning. We believe this is an important property for future benchmark design, especially for embodied or assistive systems that must justify their decisions using temporally and spatially verifiable evidence.

% \begin{table}[t] 
% \caption{Grounding Score (G) evaluation on [BenchmarkName]. Results are reported on a strict 0-5 scale.}
% \label{tab:grounding_score}
% \centering
% \small
% \setlength{\tabcolsep}{4pt}
% \resizebox{\columnwidth}{!}{% 
% \renewcommand{\arraystretch}{1.08}
% \begin{tabular}{l|c|cccc}
% \toprule
% \textbf{Method} & \textbf{Overall} & \textbf{Category I} & \textbf{Category II} & \textbf{Category III} & \textbf{Category IV} \\
% \midrule
% \midrule
% \rowcolor{gray!10}
% \multicolumn{6}{c}{\textit{Proprietary Multimodal Foundation Models}} \\
% \midrule
% Model A & -- & -- & -- & -- & -- \\
% Model B & -- & -- & -- & -- & -- \\
% Model C & -- & -- & -- & -- & -- \\
% \midrule
% \rowcolor{gray!10}
% \multicolumn{6}{c}{\textit{Open-Source Multimodal Foundation Models}} \\
% \midrule
% Model D & -- & -- & -- & -- & -- \\
% Model E & -- & -- & -- & -- & -- \\
% Model F & -- & -- & -- & -- & -- \\
% \bottomrule
% \end{tabular}%
% }
% \end{table}

\vspace{-2mm}

\section{Conclusion}

% We introduced \textbf{EgoCoT-Bench}, a fine-grained egocentric benchmark for operation-centric video understanding with structured rationale supervision and spatio-temporal evidence support. The benchmark covers egocentric grounding and perception, retrospection, predictive inference, and high-level grounded reasoning.

% Extensive experiments on a diverse set of proprietary and open-source MLLMs suggest that current progress in egocentric video understanding should be measured not only by final-answer accuracy, but also by whether a model can ground its reasoning in explicit first-person evidence. 

We present \textbf{EgoCoT-Bench}, a fine-grained benchmark for grounded, verifiable operation-centric reasoning in egocentric videos, featuring explicit spatio-temporal evidence and rationale annotations. Extensive evaluations of state-of-the-art MLLMs reveal that, despite strong answer accuracy on certain subtasks, models still struggle with evidence grounding and rationale consistency. These findings underscore the need for more reliable benchmarks and models for egocentric reasoning. We hope EgoCoT-Bench serves as a robust testbed for advancing grounded, verifiable, and temporally coherent reasoning in egocentric video understanding.

\bibliographystyle{ACM-Reference-Format}
\bibliography{bib/acmmm2026}

% %%
% %% If your work has an appendix, this is the place to put it.
% \appendix









\end{document}